# Modularizing and Assembling Cognitive Map Learners via Hyperdimensional Computing

Nathan McDonald
Air Force Research Laboratory, Information Directorate
Rome, NY, USA
Nathan.McDonald.5@us.af.mil

*Abstract*— Biological organisms must learn how to control their own bodies to achieve deliberate locomotion, that is, predict their next body position based on their current position and selected action. Such learning is goal-agnostic with respect to maximizing (minimizing) an environmental reward (penalty) signal. A cognitive map learner (CML) is a collection of three separate yet collaboratively trained artificial neural networks which learn to construct representations for the node states and edge actions of an arbitrary bidirectional graph. In so doing, a CML learns how to traverse the graph nodes; however, the CML does not learn when and why to move from one node state to another.

This work created CMLs with node states expressed as high dimensional vectors suitable for hyperdimensional computing (HDC), a form of symbolic machine learning (ML). In so doing, graph knowledge (CML) was segregated from target node selection (HDC), allowing each ML approach to be trained independently. The first approach used HDC to engineer an arbitrary number of hierarchical CMLs, where each graph node state specified target node states for the next lower level CMLs to traverse to. Second, an HDC-based stimulus-response experience model was demonstrated per CML. Because hypervectors may be in superposition with each other, multiple experience models were added together and run in parallel without any retraining. Lastly, a CML-HDC ML unit was modularized: trained with proxy symbols such that arbitrary, application-specific stimulus symbols could be operated upon without retraining either CML or HDC model. These methods provide a template for engineering heterogenous ML systems.

## I. Introduction

While classification is a popular application of artificial neural networks (ANN), there is a wide body of research showing that prediction is a critical component of cognition in biological neural networks [1, 2]. For example, when a calf is born, it spends its first several hours learning to coordinate its leg muscle movements to perform locomotion, directed movement from one place to another. By flailing about, it learns what physical states and set of actions are available to it. Whereas traditional reinforcement learning seeks to maximize (minimize) an environment reward (penalty) [3], predictive learning is task-agnostic, minimizing instead the error between a) its actual next observed state and b) its predicted next state given its current state and choice of action.

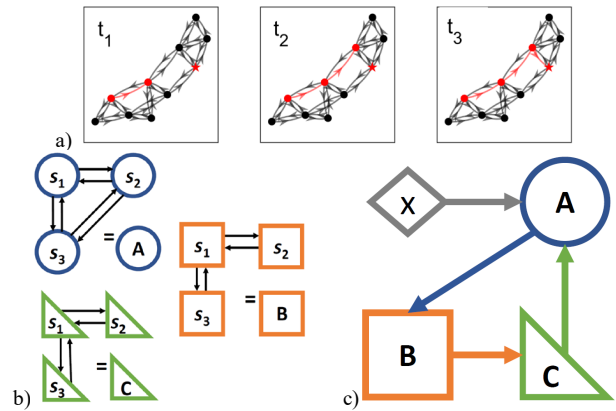

Fig. 1. a) A cognitive map learner (CML) iteratively selecting the near optimal set of edges from its current state (red node) to a user-specified target state (red star). b) By representing these node states as high dimensional vectors, independently trained CMLs may be modularized (lettered shapes) and c) arbitrarily integrated together according to hyperdimensional computing rules, enabling engineering solutions to machine learning tasks without retraining the underlying CML components.

Learning the topology of a bidirectional graph is an abstracted version of predictive learning, where each node represents an observable state and each edge an action available while in that state. A cognitive map learner (CML) was recently introduced that used three separate but collaboratively trained ANNs to construct internal representations of a) the node states, b) the node-specific edge actions, and c) edge action utility values [4]. Remarkably, though the CML was never explicitly trained for path planning, a user can specify a target node state and the CML will traverse from its current node state to the target node state along a near optimal path (fewest edges) (Fig. 1a) [4]. Note, the CML did not learn when and why to transition from one node state to another; rather, the target node state must be provided by an external source.

This work addresses the mathematics of integrating and orchestrating multiple CMLs together as a finite state machine (Fig. 1b, c), a notoriously difficult task with traditional ANNs [5]. A key challenge to integration is interfacing: making the input and output type consistent. For example, in the case of digital logic {0,1}, consistent information representation allows independently optimized logic units to be assembled together to solve problems larger than a single Boolean operation, e.g. arranging AND, OR, and XOR logic gates into a digital adder. Further, consistent interfacing permits interchangeability of components without necessarily redesigning the rest of the system.

High dimensional vectors, length $d \geq 1e3$, are herein proposed as an information representation suitable for assembling and controlling CMLs according to a hyperdimensional computing (HDC) algebra, a form of Vector Symbolic Architectures (VSA) [6, 7]. HDC has become one of the most popular non-ANN approaches to ML in recent years. Instead of learning synaptic weights values, HDC encodes learning by manipulating the similarity among a set of hypervectors, comprised of {0, 1} [8], {-1, +1} [9], or complex values [10]. Being an algebra, such learning may be explicitly expressed as equations that can be edited and reverse engineered, affording both human interpretation and intervention [11].

The contributions of this work are as follows:

- Detailed viable hypervector generation from the node state matrix of a trained CML, $d$ = 1e3, 1e4.

- Showed CMLs can train around a user-specified node state matrix, permitting engineering of arbitrarily high CML hierarchies.

- Implemented an HDC stimulus-response model to learn and decide a target state for a single CML.

- Demonstrated the addition of multiple such models into a monolithic HDC experience model to control multiple CMLs in parallel, without retraining either the HDC model or the underlying CMLs.

- Developed a method to modularize a pre-trained CML-HDC ML unit, accepting application-specific hypervector inputs without retraining either the HDC model or the CMLs.

Section II describes the construction and training of a CML and introduces the HDC algebra rules. Section III describes the methods for generating hypervectors from CML node state representations, enabling the subsequent integration of multiple, independently trained CMLs. Section IV details the integration results, followed by discussion and future applications of this research in Section V.

## II. BACKGROUND

### A. Cognitive Map Learner (CML)

The cognitive map learner is a system of three separate but collaboratively trained single-layer ANNs [4]. While CMLs may learn other tasks as well, this work focused on bidirectional graphs (Fig. 2). Each edge refers to an action permissible only between those two node states. Bidirectionality requires each action to be reversible. In this work, "node" and "state" will be used interchangeably, as will "edge" and "action."

A CML learns three fundamental things: a) internal state representations for each node ($W_q$), b) the set of available actions at each node ($W_k$), and c) the utility of each action at a given node ($W_v$). Given a graph of $n$ nodes and $e$ edges with a state representation length $d$, matrices $W_q \in \mathbb{R}^{(d,n)}$ and $W_v \in \mathbb{R}^{(d,e)}$ are both initialized as random Gaussian distributions, $\mu$ = 0, with $\sigma$ = 0.1 and $\sigma$ = 1, respectively. $W_k \in \mathbb{R}^{(e,d)}$ is initialized as a zeros matrix. $d$ is left as user defined in the original paper, and by default was set to $2n$.

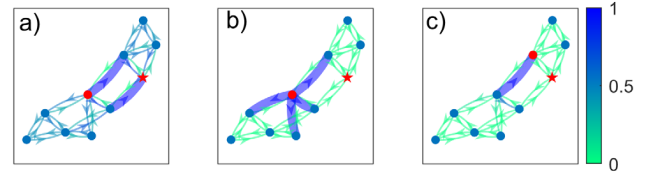

Fig. 2. Given starting node state $s_t$ (red circle) and target node state $s^*$ (red star), a CML's ANNs calculate for each edge action a) a utility value $u_t$, b) a permissibility value $g_t$ given $s_t$, and 3) a final selection value $g_t \odot u_t$. Highest valued edges have increased widths for visual clarity.

At time $t$, the CML receives an observable state vector $o_t$ which comes from the identity matrix, $I \in \mathbb{R}^{(n,n)}$, where the index matches the current node index. The internal state representation of this observation is calculated as

$$s_t = W_q\, o_t. \quad (1)$$

Critically, since $o_t$ comes from the identity matrix, $s_t$ is unambiguously the $t^{th}$ column of $W_q$, which will be exploited in the rest of this work.

The action vector $a_t$ also comes from an identity matrix, $I \in \mathbb{R}^{(e,e)}$, where each edge from the graph is uniquely indexed. The next observable node state is predicted, $\hat{s}_{t+1}$, as a function of the current state and the selected action,

$$\hat{s}_{t+1} = s_t + W_v\, a_t. \quad (2)$$

During each training epoch, all $e$ actions are performed. To encode action permissibility, a gating function $g$ is also learned,

$$g_t = f_g(W_k\, s_t), \quad (3)$$

where $f_g$ is a generalized sigmoid function

$$f_g(x) = \text{clip}\left(\frac{\sigma(x) - \sigma(0)}{\sigma(\alpha) - \sigma(0)}, 0, 1\right), \quad (4)$$

where $\sigma$ is the sigmoid function and $\alpha$ is a constant defining the saturation point of the function. During the learning phase, $\alpha$ = 1; while during testing, $\alpha$ = 0.1 [4].

To train $W_q$, $W_v$, and $W_k$, a simple delta learning rule is used [12]; whereby, the change is calculated as the difference between the actual and the predicted values multiplied by the transpose of the effecting input,

$$\Delta W_k(t) = l_k(a_t - g_t)s_t^\mathsf{T}, \quad (5)$$

$$\Delta W_v(t) = l_v(s_{t+1} - \hat{s}_{t+1})a_t^\mathsf{T}, \quad (6)$$

$$\Delta W_q(t) = l_q(\hat{s}_{t+1} - s_{t+1})o_{t+1}^\mathsf{T}, \quad (7)$$

where $l_k$, $l_v$, and $l_q$ are the respective learning rates. Here, $l_k = l_v = l_q = 0.1$. For simplicity, weight updates are summed and applied at the end of each training epoch. Regularization is then applied to preserve unit length among vectors,

$$W = \frac{W}{\|W\|^2}. \quad (8)$$

Note, $W_v$ is regularized along the $e$ axis, while $W_k$ is normalized along the $d$ axis.

To use the CML for traversing its graph, first a starting observation $o_t$ is selected by the user, setting the current state $s_t$,

(1). Next, a target observation $o*$ is selected by the user, with a target state of

$$s* = W_q\, o*. \qquad (9)$$

Second, the utility of every action is calculated by multiplying the difference between the target and current state by the transpose of the $W_v$ matrix [4] (Fig. 2a),

$$u_t = W_v^T (s* - s_t). \qquad (10)$$

Third, the gating function $g_t$, (3), calculates the indices of the permissible actions at state $s_t$ (Fig. 2b). Then the selected action is determined by multiplying $g_t$ and $u_t$ elementwise (Fig. 2c) and applying the winner take all (WTA) function to create a one-hot vector,

$$a_t = \text{WTA}(g_t \odot u_t), \qquad (11)$$

where $\odot$ denotes elementwise multiplication. Lastly, the next predicted state $\hat{s}_{t+1}$ is calculated, (2); and, by iterating over (10, 11, 2), the CML finds the (near) optimal set of actions between any starting state and target state, even though the network was never explicitly trained on such traversal tasks (Fig. 1a).

### B. Hyperdimensional Computing (HDC)

While hyperdimensional computing shares ideas about information representation found in the neural activity in the brain, it is an algebra rather than a new class of ANN. Instead of artificial neurons and synapses, HDC performs symbolic computing with hypervectors, vector of length $d \geq 1e3$. The essential metric of HDC is similarity, thus the chief concern shifts from the *location* of the dissimilar elements (e.g. most/least significant bits and error correction codes) to the *quantity* of mismatches. In so doing, every element becomes equally (in)significant for defining a particular symbol. As the length of these randomly generated vectors increases, they are effectively guaranteed to be pseudo-orthogonal [6]. Thus, if two symbols are not pseudo-orthogonal, then there must be some correlation between them. For this work, dense hypervectors comprised of uniform random {-1, +1} elements were used according to the Multiplication, Addition, and Permutation (MAP) approach [9].

Similarity between vectors is measured by their cosine similarity, their dot product divided by the product of their respective magnitudes,

$$sim_{cosine} = \frac{x \cdot y}{\|x\|\|y\|}. \qquad (12)$$

Identical vectors have a cosine similarity of 1, while pseudo-orthogonal hypervectors have a similarity close to 0. The mean similarity value of randomly generated hypervectors of length $d = 1e3$ is $0.0 \pm 0.032$ ($d = 1e4$ is $0.0 \pm 0.009$), yet the similarity range can be as high as 0.1 (0.03) (Fig. 3). Whether the standard deviation or the maximum similarity value is the appropriate noise floor threshold $\theta$ is application dependent.

The basic operations within HDC are the addition, multiplication, and permutation of hypervectors. These are elementwise operations, so no matter how many hypervectors

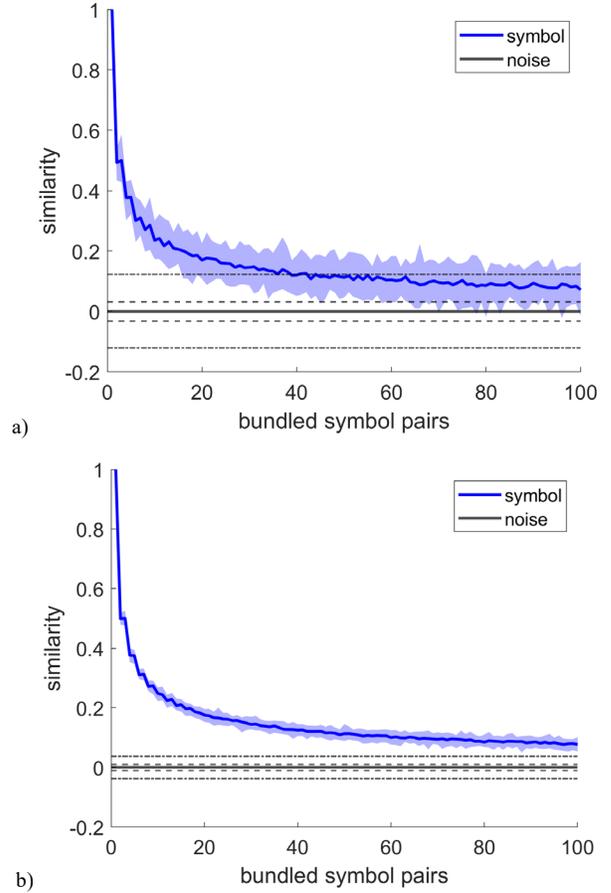

Fig. 3. For a) $d = 1e3$ and b) $d = 1e4$, noisy symbol similarity (blue) as a function of the number of bound pairs bundled together (mean enveloped by extremum). Noise floor (black) defined as similarity between random symbols (mean: solid line, standard deviation: double dash, extremum: dot-dash)

are added or multiplied together, the dimension of the resultant hypervector remains $d$. (Permutation is simply a circular shift of the indices of the vector, denoted as $\Pi$, and was not employed here.)

Hypervectors may be added (bundled) together using signed addition, clipping values back to {-1, +1},

$$s = [x + y + z]. \qquad (13)$$

When bundling an even number of hypervectors, a random hypervector $\eta$ is included to break ties. Since information is encoded along the entirety of the hypervector, the bundling operation is akin to creating a superposition of each symbol across $s$. Given then the composite hypervector $s$ and the dictionary $D$ of symbols $\{w, x, y, z\}$, one can identify (and reconstruct) the individual vectors comprising $s$,

$$sim(s, x) \approx sim(s, y) \approx sim(s, z) \gg sim(s, w). \qquad (14)$$

Multiplication (or binding) of hypervectors, denoted by $\otimes$, binds two symbols together, analogous to key-value pairing. Unlike with addition, the resultant hypervector is not similar to either of its component vectors. Here, elementwise multiplication is the binding operator, and it is a self-reversible operation.

Let $s = [w \otimes x + y \otimes z]$. To recover the hypervector bound with $w$,

$$w \otimes s = w \otimes [w \otimes x + y \otimes z] \quad (15)$$
$$= \cancel{w \otimes w} \otimes x + w \otimes y \otimes z$$
$$= x + \eta$$

where $\eta$ is a random hypervector. Since arbitrary hypervectors are nearly always pseudo-orthogonal, $w \otimes y \otimes z$ may be treated as equivalent to a random hypervector [8]. Comparing $x+\eta$ to all other symbols in dictionary $D$ reveals the greatest similarity to symbol $x$. However, because multiplication is distributive, the number of extraneous terms consolidated into $\eta$ increases rapidly with the number of symbols bound and bundled together, decreasing $sim(x, x+\eta)$ (Fig. 3). So long as the noisy symbol has similarity greater than $\theta$, complete error correction via dictionary search can be achieved.

### III. METHODS

#### A. Training Hypervector Node States

To restrict the evaluation space, this work trained CMLs on randomly generated graphs of $n = \{10, 25\}$ nodes with $e \sim 2n$ edges. A successfully trained CML a) selected the correct edge action to move between all pairs of neighboring nodes and b) successfully transitioned between 50 randomly selected start $s_t$ and target $s^*$ node states without invoking any edge action unavailable at its current state. $s_t$ and $s^*$ were at a minimum of two edges apart. The effect of state vector length $d$ on CML performance was measured as the fraction of successfully trained CMLs given a random initialization of edges, $W_q$, $W_v$, and $W_k$.

Training 50 CMLs for 500 epochs each over (5-7) for $d = 2n$ achieved a success rate of 1 for $n = 10$ (0.92 for $n = 25$) but reached 1 for both $n$ when $d = \{1e2, 1e3, 1e4\}$. For the purposes of working with HDC, hypervector lengths were fixed to $d = \{1e3, 1e4\}$ for the rest of this paper.

Trained $W_q$ matrices were dense, with a Gaussian distribution of values over [-0.1, 0.1] (Fig. 4a). Attempts to train a viable CML while forcing $W_q$ values to be $\{-1, +1\}$ failed every time (not shown). However, applying the sign function,

$$sgn(x) := \begin{cases} -1 & if\ x < 0, \\ 0 & if\ x = 0, \\ 1 & if\ x > 0, \end{cases} \quad (16)$$

to $W_q$ produced the desired a) equiprobable $\{-1, +1\}$ distribution and b) the pseudo-orthogonality of node states, $sim(sgn(s_i), sgn(s_j))$, where $i \neq j$ (Fig. 4b). These fundamental results enabled the rest of the paper's experiments.

Additional observations included a clear separability between $sim(s_i, sign(s_i)) \approx 0.8$ and $sim(s_i, sign(s_j)) \approx 0$, where $i \neq j$ (Fig. 5). This high similarity means that $sign(s_t)$ can be used directly as the target state $s^*$ when interacting with a CML, though a dictionary cleanup operation can always be used to recover the unmodified $s_t$. For notational convenience, however, all subsequent discussions of $W_q$ and $s$ in this work refer only to their signed versions unless otherwise specified.

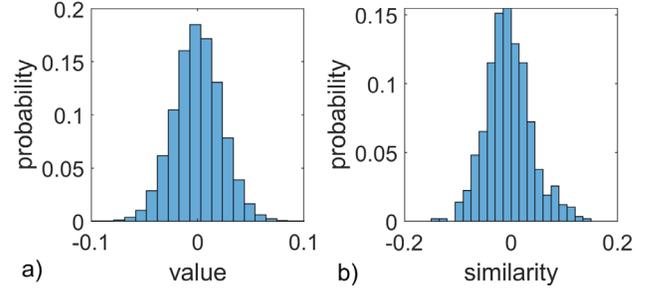

Fig. 4. Histogram of a) trained $W_q$ values and b) similarity among state hypervectors in $sgn(W_q)$ for $n = 25$, $d = 1e3$

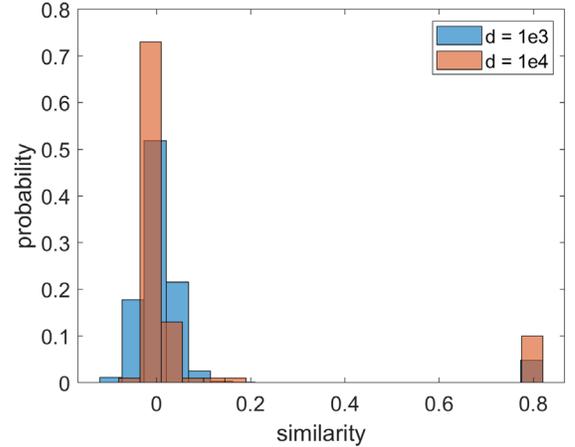

Fig. 5. Histograms of similarity between node states in $W_q$ and $sign(W_q)$ for $d = 1e3$ (blue) and $1e4$ (orange).

Lastly, the delta leaning rule could train to accommodate a predefined $W_q$. Randomly initializing $W_q$ with $\{-1, +1\}$ values followed by regularization, (8), still attained a CML training success rate of 1. A slight change from "randomly" to "arbitrarily" facilitated multiple approaches to integrating different CMLs together.

#### B. Training an HDC Experience Model

The following HDC stimulus-response experience model was introduced in [14]. HDC is used to associate input scenes, arbitrary HDC symbols, with responses, a node state of a CML (Fig. 6a). Each of $m$ inputs transmits from its own dictionary of $k$ states, denoted as $x_k^m$. Here, inputs are arbitrary hypervectors, but in practice, these input symbols may come from environmental sensors or other CML node states.

An input *scene* is comprised of a set of $m$ input states,

$$scene_i = [x_i^1 + x_i^2 + \ldots + x_i^m], \quad (17)$$

While $i$ is used among all inputs for notational convenience, it refers to an arbitrary input symbol. During training, each *scene* is paired with a target node state, creating a *scenario*,

$$scenario_i = scene_i \otimes s_i. \quad (18)$$

Finally, an experience model is created by bundling all the *scenarios* together,

$$EXP = \sum scenario_i. \quad (19)$$

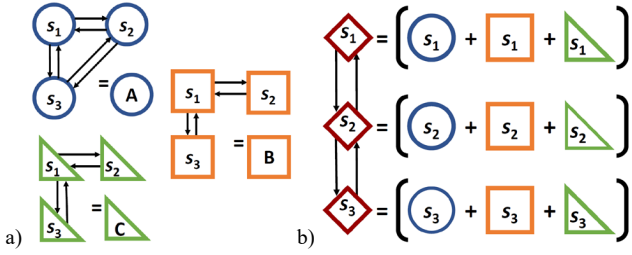

Fig. 6. a) CML *A* (blue cricles), *B* (orange squares), and *C* (green triangles) whose nodes are b) bundled togther as node states of a hierchical CML

It is good practice, to validate the *EXP* model by querying it with the training *scenes*,

$$response^{val} = scene_i \otimes EXP. \quad (20)$$

The similarity between *response*$^{val}$ and dictionary $W_q$ is measured, and the node state with the highest similarity above the noise floor threshold $\theta$ is returned as the specified target state,

$$s_i = cleanup(response, W_q, \theta). \quad (21)$$

So long as the similarity values for *response*$^{val}$ remain above the noise floor, *EXP* will return the correct target state.

Because the similarity of any *response* to its target state need only be greater than $\theta$, the deployed *EXP* model may be queried with *scenes* comprised of additional input symbols without recalculating *EXP*,

$$query = [x_i^1 + x_i^2 + \ldots + x_i^m + \eta], \quad (22)$$

$$response = query \otimes EXP, \quad (23)$$

cleanup proceeding as before, (21). For *responses* sufficiently similar (>$\theta$), the CML will accept the new target state and begins to traverse its graph. If *response* is insufficiently similar, the CML ignores it.

For measuring the *EXP* model performance, noise similarity thresholds were selected to minimize spurious CML state changes (false positives), accepting an increase in missed CML state transitions (false negatives) per *scene*. Specificity was measured as the fraction of true negatives (TN) to the total true negatives and false positives (FP),

$$specificity = \frac{TN}{TN + FP}. \quad (24)$$

Sensitivity was measured as the fraction of true positives (TP) to the total true positives and false negatives (FN)

$$sensitivity = \frac{TP}{TP + FN}. \quad (25)$$

## IV. RESULTS

For each of the following experiments, a randomly generated CML was trained and tested according to the experiment criteria. Reported values are averaged over 5 trials for each parameter setting. The number of nodes per CML was $n = \{10, 25\}$, and the hypervector length was $d = \{1e3, 1e4\}$.

### A. Heircharies of CMLs

The simplest augmentation of CMLs through hypervector node states is the construction of hierarchical CMLs. A new $W_q$

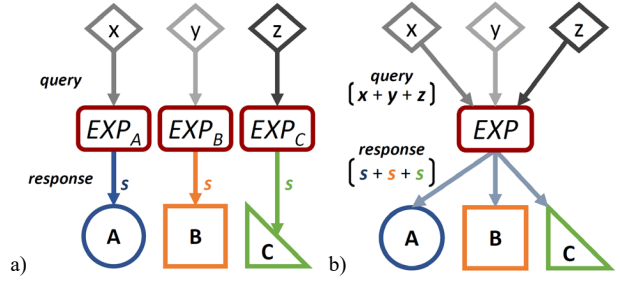

Fig. 7. a) Three independently trained CML (shapes *A*, *B*, *C*) with their respective *EXP*. b) Bundling the *EXP* allowed coslidation of both *query* and *response*.

node state matrix for *n* nodes is constructed by bundling together *n* nodes, one node state from each of *n* lower level CMLs (Fig. 6). The higher level CML trained around this predefined $W_q$. As the higher level CML traverses its bidirectional graph, each of its node states specifies the target state for each of the *n* lower CMLs, which perform the requisite cleanup, (21).

Since CMLs can be trained around random $W_q$ matrices which also produce pseudo-orthogonal hypervectors, it is sufficient to simulate 50 levels of hierarchical CMLs $n = \{10, 25\}$ using random $d = 1e4$ hypervectors. Starting with symbol $s^1$, the target symbol in level 1, the next level node state $s^2$ was created by bunding $s^1$ with *n*-1 randomly generated hypervectors, representing a CML where each node state is the sum of *n* other CML node states (Fig. 6b). More generally, the $s^{v+1}$ node state at each CML level was calculated as

$$s^{v+1} = [s^v + \sum^{n-1} \eta]. \quad (26)$$

With reference to Fig. 3, $s^1$ became pseudo-orthogonal to (irrecoverable from) higher nodes by $s^3$. However, the similarity between adjacent level nodes, $s^v$ and $s^{v+1}$, remained consistent at 0.24 ± 0.04 for $n = 10$ (0.16 ± 0.02 for $n = 25$), well above the noise floor. Since each $v^{th}$ level CML only needs to clean up received $v+1$ states with respect to its own dictionary of node states, this result shows that one can start from an arbitrary $s^v$ node and always sequentially reconstruct the lower-level nodes bundled together to create it.

### B. Monolithic HDC Experience Model

But hierarchical CMLs only push the initial target state selection algorithm further up the CML hierarchy. Using HDC for symbolic ML directly enabled learning associations between input symbols with CML target states (Fig. 7). CML *A*, *B*, and *C* received inputs *x*, *y*, and *z*, respectively, with $k = 1$-5 symbols each. Each *EXP* model learned *k* scenarios, associating each input to one target state,

$$EXP_A = \sum_{i=1}^{k} x_i \otimes s_i^A, \quad (27)$$

$$EXP_B = \sum_{i=1}^{k} y_i \otimes s_i^B, \quad (28)$$

$$EXP_C = \sum_{i=1}^{k} z_i \otimes s_i^C. \quad (29)$$

After validating each experience model, (20), they were bundled together without any retraining to create a monolithic *EXP*,

$$EXP = [EXP_A + EXP_B + EXP_C]. \quad (30)$$

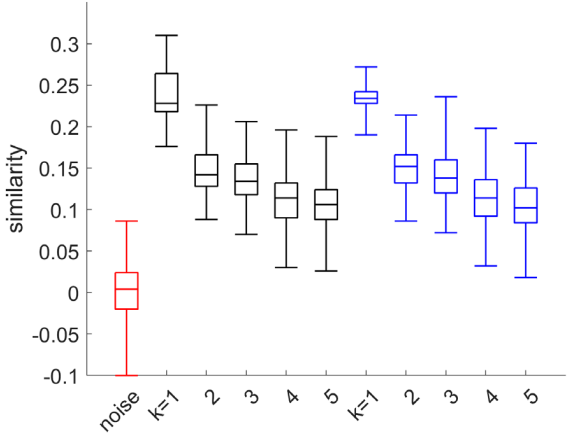

Fig. 8. Box plots of noise (red) and *response* similirity to target state $s_i$ over $3k$ learned input states for $d = 1\text{e}3$, $n = 10$ (black) and $n = 25$ (blue)

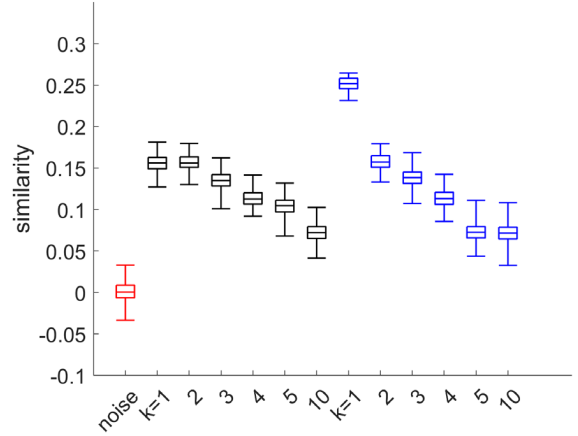

Fig. 9. Box plots of noise (red) and *response* similirity to target state $s_i$ over $3k$ learned input states for $d = 1\text{e}4$, $n = 10$ (black) and $n = 25$ (blue)

This combined *EXP* was validated with the collective $3k$ training *scenes*, classifying the resultant target state after cleanup, (20, 21). For all $n$ and $d$ combinations, sensitivity and specificity were 1 over the training *scenes* for the combined *EXP*.

To test the robustness of this bundled *EXP* approach, each input was expanded to $2k$ possible symbols, $k$ trained symbols and $k$ novel symbols. One testing cycle consisted of running through each of the $3k$ training scenes bundled with randomly selected symbols from the other two inputs. That is, each test *scene* had 3 symbols in some combination of 1-3 trained symbols with 0-2 novel symbols. This bundled *scene* produced a bundled *response* when queried of *EXP* (Fig. 7b),

$$response = [x_i + y_i + z_i] \otimes EXP \quad (31)$$
$$= [s_i^A + s_i^B + s_i^C + \eta].$$

The same *response* vector was relayed to each CML to clean up according to its respective $W_q$.

Each such testing cycle was repeated 10 times, randomly selecting the other input symbol indices each time, for $30k$ scenes for each of 5 trials (3 inputs, $k$ trained symbols, 10 cycles). The similarity of the combined *response* was measured against each CML's $W_q$. The similarity of *response* to its target node states are plotted as a function of $k$ (Fig. 8, 9), where box plot lines indicate maximum, third quartile, mean, first quartile, and minimum data values, top to bottom.

As previously noted, the noise floor was the limiting factor for the number of learned *scenarios*. Since any of the 3 CMLs might not receive a new target state per *scene*, (31), the noise floor threshold was set to the noise maximum similarity to avoid "recovering" spurious target states. Unsurprisingly, increasing $k$ (corresponding to $3k$ scenarios bundled in *EXP*) decreased the overall *response* similarity. For $d = 1\text{e}3$, the *response* similarity overlapped with the noise floor by $k = 3$ for both $n$ (Fig. 8). A threshold of $\theta = 0.08$ maintained a near prefect specificity $\geq 0.99$ for both $n$, accepting a decreased sensitivity over $k$ (Table I). For $d = 1\text{e}4$, the *response* similarity remained above the noise floor for both $n$ (Fig. 9), such that for $\theta = 0.04$, perfect sensitivity and specificity were attained. $k = 10$ was also tested with a negligible drop in sensitivity to $0.99 \pm 0.01$.

TABLE I.  MONOLITHIC *EXP* SENSITIVITY, $d = 1\text{E}3$

|   |    | k |   |   |   |   |
|---|----|---|---|---|---|---|
|   |    | 1 | 2 | 3 | 4 | 5 |
| n | 10 | 1 | 1 | $0.97 \pm 0.01$ | $0.89 \pm 0.03$ | $0.86 \pm 0.04$ |
|   | 25 | 1 | $0.99 \pm 0.01$ | $0.95 \pm 0.03$ | $0.88 \pm 0.02$ | $0.82 \pm 0.02$ |

### C. CML Interfaces with Proxy Symbols

While an *EXP* is resilient to a certain amount of extraneous sensor input [13], if the set of trained input symbols changes, then *EXP* will need to be retrained on the new set of symbols (Fig. 1c, 10a). Instead of relearning specific *scenarios* every time an input changes, one can use HDC to encode *scenarios* with proxy input symbols and map application-specific inputs to these proxy symbols. For example, in Fig. 10a, CML *A* has inputs $x$ and $s^D$; however, HDC allows for creating a generalized *EXP* with two inputs (Fig. 10b).

Template scenes $scene^0$ were created from $m = 1$-8 inputs with $k = 1$-8 proxy states each,

$$scene_i^0 = [p_i^1 + p_i^2 + \ldots + p_i^m]. \quad (32)$$

These *scenes* were bound to $k$ target node states,

$$EXP = \sum_{i=1}^{k} scene_i^0 \otimes s_i. \quad (33)$$

To interact with the CML, one requires knowledge of the proxy symbols $p$, $scene^0$ templates, and $W_q$ node states. A novel symbol $x$ was then bound with each of the $km$ learned proxy input symbols. These bound pairs were then bundled together to create a *map*,

$$map = \sum_{i=1}^{k}[x_i^1 \otimes p_i^1 + x_i^2 \otimes p_i^2 + \ldots + x_i^m \otimes p_i^m]. \quad (34)$$

During testing, the application specific *scene* was multiplied by the *map*, resulting in an approximate version of one of the $scene^0$ templates,

$$query = scene \otimes map$$
$$= [x_i^1 + x_i^2 + \ldots + x_i^m] \otimes map$$
$$= [\varkappa_i^1 \otimes x_i^1 \otimes p_i^1 + \varkappa_i^2 \otimes x_i^2 \otimes p_i^2 + \ldots + \varkappa_i^m \otimes x_i^m \otimes p_i^m + \eta]$$
$$= [p_i^1 + p_i^2 + \ldots + p_i^m + \eta]. \quad (35)$$

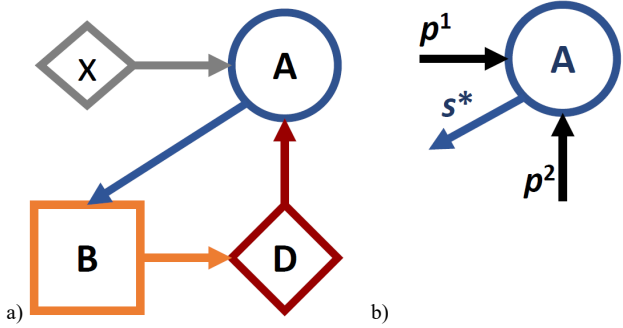

Fig. 10. a) A CML integrated with HDC allows for module replacement (see Fig. 1c) without complete retraining of the entire ML system. b) The *EXP* model for CML *A* can be trained with task-agnostic proxy symbols, which are then mapped to application specific inputs $x$ and $s^D$.

Note, since multiplication is associative, $\eta$ consolidated a large number of meaningless terms, e.g. $x_i^2 \otimes x_i^1 \otimes p_i^1$ occurs in (35). These noise terms caused significant symbol degradation (Fig. 3), so the *scene*$^0$ templates were critical to clean up the *query* before calculating the *response* per usual,

$$scene_i^0 = cleanup(query, scene^0, \theta). \quad (36)$$

Because the user has access to the proxy symbols $p$ and $W_q$, additional *scene*$^0$ and *scenario*$^0$ templates may be created as required. Lastly, since every *query* was expected to recover a target state, the standard deviation of the noise floor was used for setting the threshold instead of the maximum range.

A complete *map* returned a *query* = *map*⊗*scene* whose maximum similarity was to the correct *scene*$^0$ template. The consistency of generating such a map with randomly generated hypervectors (representing arbitrary application-specific inputs) was measured as the fraction of trials required to generate 10 complete maps for the trained *EXP* for $d = 1e3$ (Fig. 11a) and $d = 1e4$ (Fig. 11b). ($n = 25$ for all experiments, though only $k$ node states were used.) If 10 *maps* were not generated within 300 attempts, the consistency was marked 0. For $d = 1e3$, while the consistency of successful *map* generation decreased rapidly, complete *maps* were found for $mk \leq 35$ symbols. Using $d = 1e4$, the consistency decreased more slowly and *maps* were found for all $m$ and $k$ save for $mk = 64$.

While there was a high success rate for generating *maps*, especially for $d = 1e4$, applying the mapped *query*, (35), directly to *EXP* resulted in a rapid degradation in the *response* similarity and thus target node sensitivity (once beyond the trivial $m \leq 2$ or $k \leq 2$) (Fig. 12). After performing a *query* cleanup step, (36), each successful *map* in Fig. 11 demonstrated a perfect *response* sensitivity of 1.

## V. DISCUSSION

The segregation of knowledge among the ANNs of a CML permitted the precise extraction (or defining) of node state representations. Expressing CML node states as hypervectors allowed HDC to be used to assemble and concurrently operate independently trained CMLs without subsequent retraining. Since the noise floor was the limiting factor in these experiments, adopting a hypervector length of $d \geq 1e3$, ensured access to the greatest $[\theta, 1]$ similarity range.

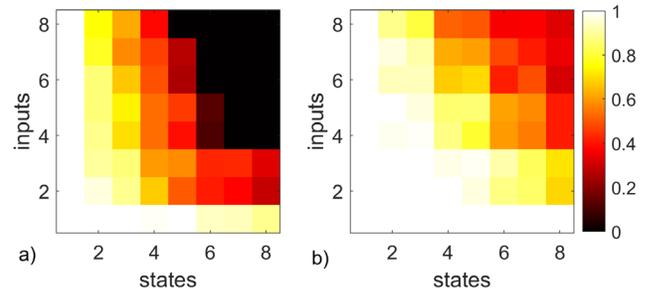

Fig. 11. Probability of generating successful map between arbitrary input vectors and proxy vectors for a) $d = 1e3$ and b) $d = 1e4$.

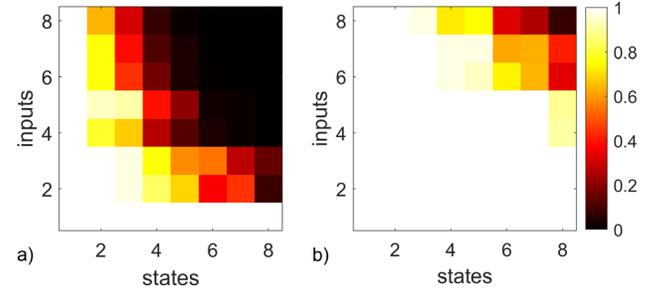

Fig. 12. Accuracy using *query* = *map*⊗*scene* without cleanup to invoke *response* for a) $d = 1e3$ and b) $d = 1e4$.

Life-long learning refers to the ability to sequentially accommodate new learning or behaviors with minimal degradation of prior knowledge [15]. CMLs algorithmically enable the segregation of graph traversal from the target node decisions mechanism, allowing engineering of more complicated ML systems. HDC permits some variation in input *scenes* to incur the same *response*, else new *scenarios* may be trivially added to the current *EXP* model. Alternatively, even higher-level ML algorithms might be used to decide the most appropriate *EXP* to use among several available models based on environment or historical context.

"Plug & play" ML for robotics is one expected application for these CML-HDC modules. Each node in a graph may represent a position of an arm and the graph itself define a particular behavior, e.g. grasping or walking. Training an CML-HDC *EXP* input-response model per appendage, one may add (or remove) the appendage to a baseline robot and operate it without any additional retraining (provided target similarities remain above the noise floor).

Tasks such as quadrupedal walking, however, require cooperation among (potentially asymmetric) appendages. Bundling node states from each of the four CMLs is a simple way to create a hieratical CML whose nodes define the respective states of all four leg CMLs. For example, each higher level CML nodes might represented a single frame in Eadweard Muybridge's photographs of animal locomotion [16].

This application also illustrates the advantages of CML interfacing with proxy symbols. Even assuming four identical legs, the input symbols may differ over the front-back, left-right leg configurations. A single CML-HDC model could be trained over the expected range of inputs and outputs, then each individual leg uniquely mapped to the same *EXP* model.

A major challenge to the greater adoption of HDC is the need for algorithms to map real-valued sensor data to hypervector symbols [17, 18]. Yet since ANNs have a rich history in classification: mapping raw sensor data to arbitrary class labels, there is recent work training ANNs as task-agnostic feature extractors [5, 19] then mapping these sparse feature vectors to arbitrary hypervector symbols for subsequent HDC computation [20, 21, 22]. This approach, in effect, turns ANNs themselves into modular ML components, functioning as the ML equivalent of analog to digital (A2D) converters.

Lastly, the CML algorithm operates over real-valued neural networks; but the illustrative biology examples described are based on spiking neural networks (SNN). Future research will focus on implementing an SNN version of CMLs based on resonate-and-fire (RF) neurons [23]. These types of SNNs encode information based on the time a neuron spikes within period $\tau$ as opposed to rate encoding, measuring the number of spikes within a time window. Importantly, a spike at time $t$ with respect to a local oscillator of period $\tau$ may be expressed as a complex valued phasor (or phase vector). RF neurons therefore also facilitate HDC interfacing via Holographic Reduced Representations (HRR), based on complex phasors [10].

## VI. Conclusion

Cognitive map learners (CML) are a collection of separate yet collaboratively trained ANNs, which learn to traverse a bidirectional graph. This work created CMLs with graph node states expressed as high dimensional vectors, with the mathematical properties required for hyperdimensional computing (HDC), a form of symbolic machine learning. Expressing CML node states as hypervectors allowed HDC to be used to assemble and concurrently operate independently trained CMLs without subsequent retraining. This work constructed an arbitrary number of hierarchical CMLs, where each graph node state specified the target node states for the next lower level CMLs to traverse to. An HDC-based stimulus-response experience model was created for each CML then bundled together allowing parallel operation again without any retraining. Finally, a mapping algorithm was developed to enable HDC model training on proxy symbols, which were then mapped to application-specific input symbols, thus creating composable CML-HDC ML modules.

## Acknowledgment

Any opinions, findings and conclusions, or recommendations expressed in this material are those of the authors, and do not necessarily reflect the views of the US Government, the Department of Defense, or the Air Force Research Lab.

DISTRIBUTION STATEMENT A. Approved for Public Release; Distribution Unlimited: AFRL-2023-0658